# TornadoNet: Real-Time Building Damage Detection with Ordinal Supervision

Robinson Umeike[1], Cuong Pham[4], Ryan Hausen[3], Thang Dao[1], Shane Crawford[1], Tanya Brown-Giammanco[5], Gerard Lemson[3], John van de Lindt[2], Blythe Johnston[2], Arik Mitschang[3], Trung Do[4]

[1] The University of Alabama  [2] Colorado State University  [3] Johns Hopkins University  [4] University of South Alabama  [5] National Institute of Standards and Technology

## 1. Abstract

We present TornadoNet, a comprehensive benchmark for automated street-level building damage assessment that evaluates how modern real-time object detection architectures and ordinal-aware supervision strategies perform under realistic post-disaster conditions. TornadoNet provides the first controlled benchmark demonstrating how architectural design and loss formulation jointly influence multi-level post-disaster damage detection from street-view imagery. The benchmark delivers both methodological insights for ordinal object detection and immediately deployable tools for accelerating disaster response efforts. Using 3,333 high-resolution geotagged images and 8,890 annotated building instances collected following the 2021 Midwest tornado outbreak, we systematically compare convolutional neural network (CNN)–based detectors from the YOLO family against transformer-based models (RT-DETR) for multi-level damage detection. All models are trained and evaluated under standardized protocols using a five-level damage classification framework based on damage states as defined in the IN-CORE platform, validated through expert cross-annotation. Baseline experiments reveal complementary architectural strengths. CNN-based YOLO models achieve the highest detection accuracy and throughput, with larger variants reaching up to 46.05% mAP@0.5 at 66–276 FPS on A100 GPUs, demonstrating real-time inference capability. Transformer-based RT-DETR models exhibit stronger ordinal consistency, achieving 88.13% Ordinal Top-1 Accuracy and a lower MAOE of 0.65, indicating more reliable severity grading despite lower baseline mAP. To align supervision with the ordered nature of damage severity, we introduce soft ordinal classification targets and evaluate explicit ordinal-distance penalties. Results demonstrate a clear architecture-dependent response. RT-DETR trained with calibrated ordinal supervision ($\psi = 0.5, K = 1$) achieves 44.70% mAP@0.5, a 4.8 percentage-point improvement over baseline, with further gains in ordinal metrics (91.15% Ordinal Top-1 Accuracy, MAOE = 0.56). Anchor-free YOLO detectors show limited or no benefit from ordinal supervision, while excessive smoothing ($\psi \geq 1.0$) leads to degenerate solutions that exploit class imbalance across architectures. These findings establish that ordinal-aware supervision can improve damage severity estimation when aligned with detector architecture. **Model & Data**: https://github.com/crumeike/TornadoNet

## 2. Introduction

### 2.1. Problem Statement

The efficacy of humanitarian aid and disaster response is fundamentally contingent on the rapid and accurate assessment of structural damage, particularly within the critical hours immediately following a catastrophic event. This temporal constraint presents a significant challenge for traditional damage assessment methods [1–5], which rely predominantly on manual field surveys conducted by trained personnel. Although these conventional approaches can provide detailed information, they are inherently slow, labor-intensive, and often expose personnel to hazardous conditions. Their effectiveness is further limited by compromised infrastructure, restricting the ability to obtain a timely and comprehensive view of affected regions when such information is most urgently needed. These operational constraints create a critical information bottleneck that directly impacts emergency response, impeding the strategic allocation of resources, delaying search and rescue operations, and can ultimately exacerbate human suffering and economic loss.

Beyond delays in response, the quality and utility of data produced by manual assessment are also constrained. Damage information is frequently aggregated at coarse geographic scales, such as counties or municipalities, obscuring the building-level detail required to accurately identify severely impacted structures and prioritize interventions. Moreover, manual assessments are susceptible to cognitive biases, including reference dependence and recall errors, which can reduce objectivity and consistency across survey teams and time periods. Traditional post-tornado damage assessment methodologies, as documented in studies following the 2000 Moore, Oklahoma tornado [6], the devastating 2011 Tuscaloosa and Joplin tornadoes [7,8], and the 2013 Moore, Oklahoma event [9], have evolved considerably but continue to face these fundamental limitations. While recent initiatives such as the StEER (Structural Extreme Event Reconnaissance) Network and the DesignSafe-CI repository have advanced data collection standardization and accessibility [10,11], the underlying manual nature of these assessments still constrains their scalability and consistency. This lack of assessment standardization limits the comparability of assessments and complicates efforts to integrate damage data into broader recovery planning and resilience modeling frameworks.

The manual process of post-disaster data collection further limits the scale and frequency at which high-quality damage datasets can be produced. While extreme hazard events occur regularly worldwide, only a small fraction is documented with sufficiently detailed, large-scale datasets to support systematic analysis. This data scarcity limits opportunities for comparative research on hazard impacts and infrastructure performance. As a result, critical knowledge of key features or failure patterns, which could inform more resilient engineering designs and building codes, remains underexplored. There is, therefore, an urgent need for standardized, objective, and realistic assessment approaches capable of producing reliable, building-level damage information to support both immediate response and longer-term risk reduction efforts.

### 2.2. Motivation

The significant limitations of manual damage assessment methods necessitate a shift toward automated solutions, and recent advances in computer vision offer a compelling path forward for post-disaster assessment. The increasing availability of high spatial resolution satellite imagery and other data collection platforms, such as those deployed on Unmanned Aerial Vehicles (UAVs) and vehicles, has created an unprecedented wealth of geospatial data [3,12–14]. When coupled with advanced computer vision techniques, these data streams can be harnessed to automate post-disaster damage assessment (PDA) at a scale and speed unattainable by human teams.

Early progress in automated PDA was driven by deep convolutional neural networks (CNNs), which demonstrated strong performance in image classification and segmentation tasks. However, many early object detection frameworks were multi-stage and computationally intensive, limiting their suitability for time-critical disaster response. This motivated the development of single-stage detectors such as YOLO

(You Only Look Once) and SSD (Single Shot MultiBox Detector), which significantly improved inference speed while maintaining competitive detection accuracy. These architectures enabled near real-time processing, making automated PDA a practical possibility for large-scale disaster scenarios [15,16].

Despite their efficiency, some CNN-based detectors rely on handcrafted components such as anchor boxes and other post-processing techniques such as non-maximum suppression (NMS), which may perform sub-optimally in cluttered and visually complex post-disaster environments [16]. More recently, transformer-based detectors have emerged as an alternative paradigm, leveraging self-attention mechanisms to model global image context. Architectures such as RT-DETR (Real-Time Detection Transformer) provide fully end-to-end detection pipelines that reduce reliance on heuristic components and offer a fundamentally different approach to object localization and classification [17].

Given these distinct architectural paradigms, it remains unclear how different real-time object detection frameworks behave when applied to the visually complex and semantically ordered problem of post-disaster damage assessment using street-view imagery. In addition to detecting buildings accurately, effective damage assessment requires distinguishing between adjacent levels of damage severity, where misclassifications are not equally consequential. Most existing detectors treat damage categories as independent classes, ignoring the inherent ordinal structure of damage progression [18]. This motivates not only a systematic comparison of modern CNN-based and transformer-based detection architectures, but also an investigation into whether incorporating ordinal-aware supervision can better align model predictions with the graded nature of structural damage. Accordingly, this work examines both architectural trade-offs and loss design choices to better understand their combined impact on detection performance and damage severity estimation under real-world disaster conditions.

## 2.3. Research Objectives

This research addresses critical gaps in post-disaster damage assessment by conducting a systematic comparison of state-of-the-art CNN and transformer-based object detection architectures for street-view building damage detection. The primary objective is to understand how architectural choices and classification supervision strategies influence detection accuracy and ordinal damage recognition in real-world post-disaster imagery.

Specifically, this study investigates the following core questions: (1) how do modern CNN-based detectors, including anchor-free YOLO variants, and transformer-based detectors (RT-DETR) compare in terms of detection accuracy and reliability when applied to street-view post-disaster imagery; (2) does incorporating soft ordinal classification targets, which encode the ordered structure of damage severity, improves detection performance relative to the standard one-hot classification used in contemporary object detectors; (3) does explicitly penalizing ordinal distance between predicted and true damage states provides additional benefit beyond soft ordinal supervision, or does it instead degrade detection performance; (4) how model performance varies across damage severity levels defined by the IN-CORE(The Interdependent Networked Community Resilience Modeling Environment) classification system [14,19]; and (5) whether the effects of ordinal-aware supervision are consistent across detector families and model scales, including multiple YOLO variants and a transformer-based RT-DETR architecture.

Our approach leverages imagery collected from the 2021 Midwest U.S. tornado events using vehicle-mounted 360° cameras [3], providing a high-quality street-view dataset for training, and evaluating object detection models under realistic disaster conditions. While prior studies in automated AI disaster assessment have demonstrated the value of ground-level imagery for damage assessment [20], most existing work has focused on satellite or aerial perspectives [13,21–23]. As a result, the comparative behavior of modern real-time detection architectures on street-view data, particularly for multi-level damage classification, remains underexplored.

By adopting standardized training protocols, fixed input resolutions, and consistent evaluation metrics, this study enables a controlled comparison across architectural paradigms and loss formulations. Beyond conventional detection metrics, we incorporate ordinal-aware evaluation measures to quantify not only detection accuracy but also the severity of misclassification errors, distinguishing minor off-by-one mistakes from more consequential misclassifications.

The contributions of this work include: (1) a comprehensive benchmark of CNN and transformer-based object detection models for street-view tornado damage assessment; (2) two ordinal-aware loss formulations, including Gaussian soft classification targets and an explicit ordinal-distance penalty, both adapted to be compatible with each detector's native loss (BCE for YOLO, focal loss for RT-DETR). (3) the introduction of Ordinal Top-k Accuracy, adapted specifically for the damage detection setting where matched detections are evaluated for severity coherence, not just class correctness. (4) an empirical evaluation of soft ordinal classification targets for multi-level damage detection; (5) a critical analysis of explicit ordinal-distance penalties and their impact on detection performance; and (6) the release of TornadoNet, a curated set of trained models weights, training code, and data to support future research. Together, these contributions advance the understanding of how ordinal structure can be incorporated into object detection for post-disaster damage assessment while clarifying its practical benefits and limitations.

## 2.4. Paper Organization

This paper is organized as follows. Section 2 reviews related work in post-disaster damage assessment, object detection architectures, and existing comparative studies. Section 3 presents our methodology, including the dataset description, model architectures, and experimental setup. Section 4 details our experimental results and comparative analysis. Section 5 discusses the implications of our findings for both research and response applications. Section 6 concludes with a summary of contributions and directions for future work. We also describe the release of TornadoNet, our collection of fine-tuned models and benchmarking results, to support continued research and development in this critical application domain.

## 3. Related Work

Building on the growing need for rapid and reliable tornado damage assessment, several frameworks have been developed to classify damage severity. The most widely used is the Enhanced Fujita (EF) Scale, which rates tornado intensity based on estimated wind speeds and observed damage indicators [24,25]. For more granular, structure-level assessment, the Hazards U.S. Multi-Hazard (HAZUS-MH) methodology classifies buildings into discrete damage states based on observed component failures [26]. This framework has been adapted for different building archetypes and integrated into platforms such as IN-CORE to support community resilience modeling and risk analysis [19,27–30]. In parallel, several studies have adopted custom damage classification schemes, ranging from binary categories such as damaged versus undamaged or collapsed versus non-collapsed, to more detailed multi-class scales tailored to specific datasets or hazards [22,31–34].

A central challenge lies in applying these detailed classification frameworks rapidly and at scale, a task for which computer vision and deep learning have emerged as powerful solutions. In recent years, a wide range of deep learning models have been applied to automated damage detection and classification using satellite imagery, unmanned aerial systems (UAS), and street-view data [20,22,31–36]. As summarized in Table 1, prior studies have explored architectures, including ResNet, YOLO, UNet, DenseNet, MSNet, Support Vector Machines (SVM), and various custom CNN-based models. These efforts have been supported by publicly available datasets that enable supervised learning and benchmarking of disaster damage models [12,13,37]. In addition, cyberinfrastructure platforms such as IN-CORE [19] and DesignSafe [11] have facilitated data sharing, analysis, and standardization across the disaster research community.

These developments highlight the growing potential of image-based approaches for tornado damage assessment. However, a notable gap remains in the systematic evaluation of modern real-time object detection architectures for multi-level damage classification using street-level imagery. In particular, existing studies have largely focused on aerial or satellite perspectives, or on single model families, limiting insight into how contemporary CNN-based and transformer-based detectors compare under consistent training and evaluation conditions [21,31,36,38]. This study addresses this gap by providing a controlled comparison of state-of-the-art object detection architectures for street-view tornado damage assessment, with an emphasis on standardized damage labeling and ordinal severity interpretation.

Table 1. Imagery-Based Tornado Damage Detection Studies.

| Task | Framework | Classification scheme | Model | Data Source | Dataset Statistics | Tornado events | Summary of Results | Reference |
|---|---|---|---|---|---|---|---|---|
| Building-level damage | Custom | Binary (damaged/undamaged) | ResNet-18 (pre-trained on ImageNet-1000) | Satellite imagery | 37,350 in total for 11 events (including 3 tornado events) | Joplin, Moore, and Tuscaloosa–Birmingham Tornadoes (USA) | For Joplin Tornado: Accuracy: 86.7% Precision: 84.7% Recall: 85.4% F1 Score: 85% | Kim et al., 2022 [32] |
| Building-level damage | Custom | Damage 1, Damage 2, Damage 3, and Damage 4 | Faster RCNN, Inception v2, ResNet-50, ResNet-101, and Inception-ResNet-v2 | Camera, Street-view imagery | 4286 in total for training and cross-validation. | Taiwan earthquake (2016) and other earthquake events | Inception ResNet v2 achieved the highest mean AP at: 78.8% | Ghosh et al [18] |
| Roof damage | Custom (quantified % area of roof damage) | Binary (damaged/not damaged) | SVM | Satellite imagery and Aerial imagery | Aerial image data of 200m×1,500 m coverage area | Saroma Tornado (Japan) | Accuracy: 90% Correlation Factor: 0.75 (Satellite imagery) and 0.8 (Aerial Imagery) | Radhika et al., 2018 [33] |
| Building-level damage | Custom | Multi (Undamaged, DS1, DS2, DS3, DS4) | YOLO11 | Street-view imagery | 2,635 images | 2021 Midwest Tornado (USA) | Accuracy: 60.83% | Umeike et al., 2024 [20] |
| Treefall | Custom | Binary (damaged/undamaged) | UNet (with ResNet-34 backbone) | UAS imagery | 601 UAS images 6,328 imagery tiles (512×512 px) hand-labeled | Franklin County Tornado (USA) | Accuracy: 89.3% F1: 87.2% Precision: 85.7% Recall: 88.9% | Carani & Pingel, 2023 [31] |

| | | | | | | damage polygons | | Kappa: 78.0% | |
|---|---|---|---|---|---|---|---|---|---|
| Building-level damage | Custom | Multi (Slight, Severe, Debris) | MSNet | Aerial videos | 1,030 images | | Not mentioned | AP: 32.7 AP@0.5: 28.8 Bounding Box AP: 38.7 | Zhu et al., 2021 [22] |
| Building-level damage | Custom | Binary (collapsed/non-collapsed) | UNet-VGG16 | Satellite imagery | 9,008 sub-images (128×128 px); 5,021 after removing negative samples | | Kaiyuan Tornado (China) | mF1:93.85% mIoU: 89.25% mPA:94.1% | Xiong et al., 2022 [34] |
| Building-level damage | EF-Scale | Binary (damage/no damaged) and Multi (no damage, EF-0, EF-1, EF-2, EF-3) | DenseNet-161 and ResNeXt-101 | UAS imagery | 3,996 images | | Eureka, Kansas Tornado (USA) | Highest accuracy: - Binary: 84.8% (DenseNet-161) - Multi-class: 81.5% (ResNeXt-1010) | Chen et al., 2021 [36] |
| Building-level damage | IN-CORE | Multi (DS0-1, DS2, DS3, DS4) | Custom CNN | Satellite imagery | 56,000 for training 1,100 for testing | | Joplin Tornado (USA) | Accuracy: 88% Precision: 74% Recall: 76% F1 Score: 75% | Braik & Koliou, 2025 [35] |

## 4. Methodology

### 4.1. Dataset Description

This study utilizes a unique dataset of street-view imagery collected following the severe tornado outbreak of December 10–11, 2021, which caused widespread damage across nine U.S. states [14]. Data collection was conducted as part of a longitudinal field study organized by the Center of Excellence for Risk-Based Community Resilience Planning (CoE), with the objective of supporting empirical validation of the IN-CORE modeling platform [3,39].

Imagery was acquired using a vehicle-mounted system equipped with high-resolution GoPro cameras configured to capture 360° panoramic video. Data collection occurred approximately three weeks after the event, balancing the need to capture post-disaster conditions with accessibility constraints. Survey routes were designed to prioritize heavily impacted areas, informed by post-event aerial imagery, preliminary damage reports and social vulnerability indices, to ensure coverage of diverse structural damage and community contexts. This ground-level data collection approach was selected to capture architectural and structural damage to vertical building faces, such as facade failures and broken windows, that are often obscured or not visible in satellite or aerial imagery.

The raw video footage was processed using a tool to automatically extract individual building-centered image frames by geospatially aligning the vehicle trajectory with building centroids. This post-processing technique required downsampling the 5.6k resolution, panoramic video data into 4k resolution, and equirectangular images [40]. Following extraction, a manual quality control process was applied to remove images that were severely compromised by occlusions, poor lighting, motion blur, or insufficient structural visibility, while retaining challenging but informative cases representative of real-world post-disaster conditions. The final curated dataset consists of 3,333 high-resolution, geotagged street-view images and a total of 8,890 building instances.

### 4.2. Data Annotation and Splitting

Each image in the curated dataset was manually annotated by trained researchers. Bounding boxes were drawn around individual buildings, and each instance was labeled according to the IN-CORE five-level damage classification system, described in more detail in Section 3.2: DS0 (Undamaged), DS1 (Slight), DS2 (Moderate), DS3 (Extensive), and DS4 (Complete). This multi-class labeling scheme captures the ordinal nature of damage severity and enables fine-grained damage discrimination.

The annotated dataset comprises was partitioned into training, validation, and test sets following a 75%:15%:15% split respectively. The training set contains 6,184 instances, the validation set includes 1342 instances for hyperparameter tuning, and the test set consists of 1,364 instances used for final performance evaluation. The class distribution and bounding box characteristics across these splits are summarized in Figure 1.

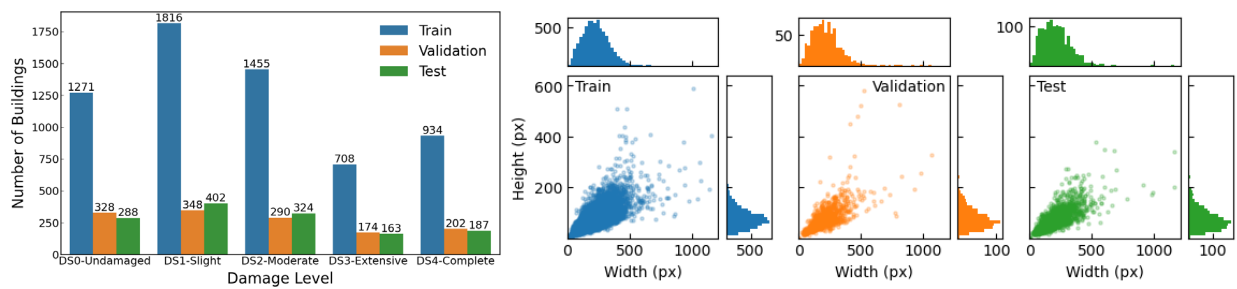

**Figure 1.** TornadoNet dataset composition. (Left) Building counts per damage class (DS0–DS4) across training (75%), validation (15%), and test (15%) splits, showing class imbalance toward lower severity levels. (Right) Bounding box dimensions (width vs. height in pixels) for each split with marginal frequency distributions.

### 4.3. Damage Classification Framework

Building damage in the dataset was annotated using the IN-CORE five-level damage state classification: DS0 (Undamaged), DS1 (Slight), DS2 (Moderate), DS3 (Extensive), and DS4 (Complete) [19]. These categories represent increasing levels of damage severity, ranging from no visible impact to complete structural failure. Each building instance was manually labeled through visual inspection by trained annotators following standardized IN-CORE guidelines [3,19].

Damage state assignments were informed by archetype-specific indicators defined for 19 structural archetypes (T1–T19) within the IN-CORE framework [19]. The majority of buildings in this dataset fall within Archetypes T1–T5, which correspond to residential wood-frame structures with varying plan dimensions and roof configurations. For these archetypes, damage states were determined using a structured decision matrix that incorporates observable failures in roof covering, windows and doors, roof sheathing, and roof-to-wall connections. The decision criteria for Archetypes T1–T5 are summarized in Table 2.

To enhance labeling reliability, all annotations underwent a secondary expert review to verify consistency and resolve discrepancies [3]. This two-stage validation process was implemented to improve annotation accuracy and reduce subjectivity. Figure 2 presents representative annotated examples across the five damage states, including both clear views and challenging conditions such as occlusions, variable lighting, long viewing distances, and image distortion.

Table 2. Damage state decision matrix for archetypes T1–T5 based on IN-CORE classification criteria.

| Element | DS1 - Slight | DS2 - Moderate | DS3 - Extensive | DS4 - Complete |
|---|---|---|---|---|
| **Roof Covering** | 2-15% of Roof Covering Damaged | 15-50% of Roof Covering Damage | More than 50% of Roof Covering Damaged | More than 50% of Roof Covering Damaged (typically) |
| | *AND/OR* | *AND/OR* | *AND/OR* | *AND/OR* |
| **Window/Door** | 1 window or door failure | 2 or 3 windows/doors failed | More than 3 windows/doors failed | More than 3 windows/doors failed (typically) |
| | *AND/OR* | *AND/OR* | *AND/OR* | *AND/OR* |
| **Roof Sheathing** | No Roof Sheathing Failure | 1-3 sections of roof sheathing failed | More than 3 sections AND less than 35% of roof sheathing failed | More than 35% of roof sheathing failed |
| | *AND/OR* | *AND/OR* | *AND/OR* | *AND/OR* |
| **Roof-to-Wall Connection** | No Roof-to-Wall Connection Failure | No Roof-to-Wall Connection Failure | No Roof-to-Wall Connection Failure | Roof-to-Wall Connection Failure |

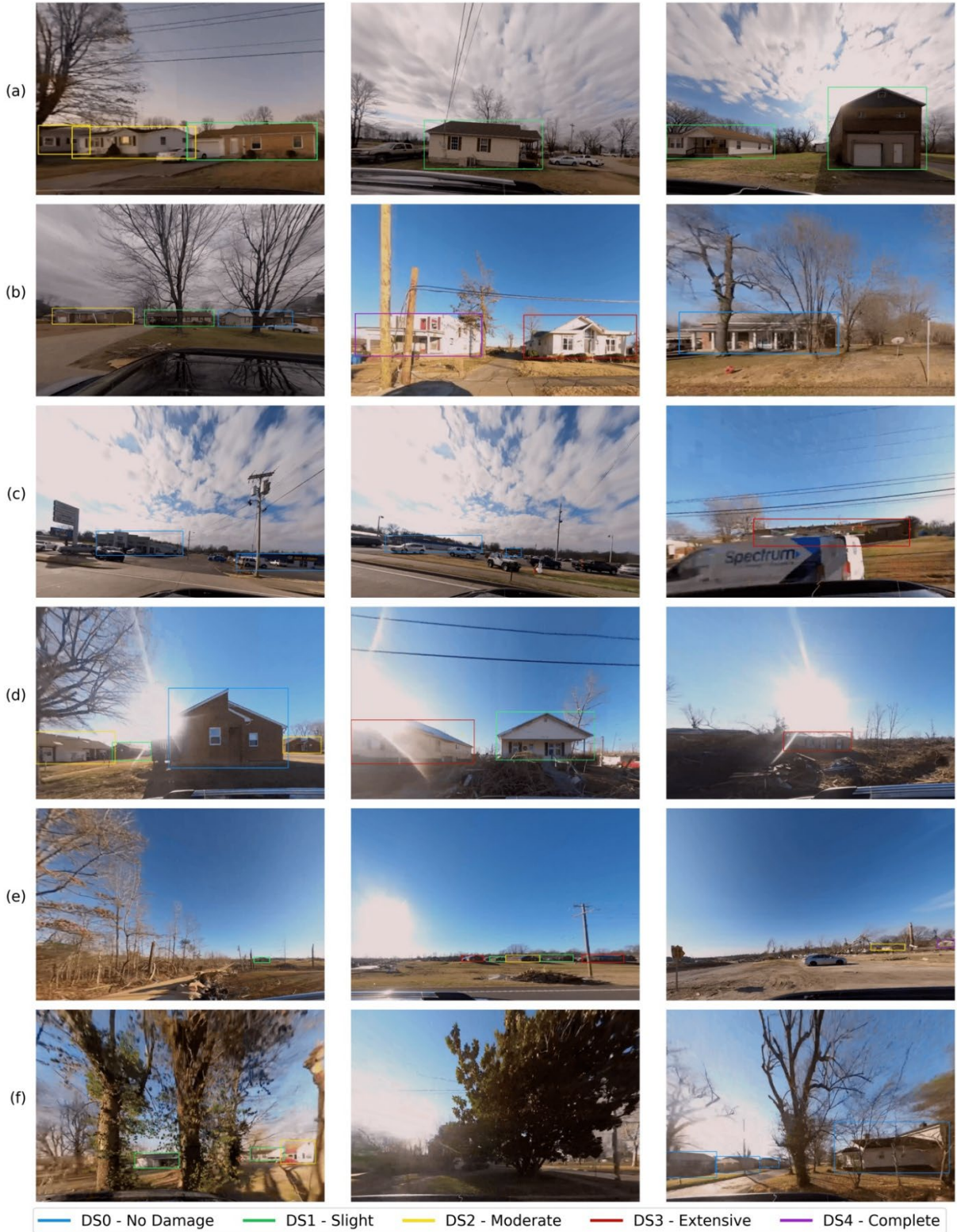

**Figure 2.** Representative annotated images showing building damage states (DS0 to DS4) in the TornadoNet dataset, under varying conditions: (a) clear and high-quality view, (b) building partially obscured by trees or poles, (c) building partially obscured by parked vehicles, (d) lighting issues such as strong sunlight or low light, (e) building located far from the camera, (f) blurry or warped imagery.

## 4.4. Object Detection Architectures

This study evaluates a diverse set of real-time object detection architectures spanning convolutional and transformer-based paradigms. All models are trained and evaluated under consistent conditions to isolate the effects of architectural design and classification supervision.

**CNN-based Detectors**. We considered multiple generations and scales of the YOLO (You Only Look Once) family, including YOLOv8, and YOLO11, each evaluated at the nano (n), large (l), and extra-large (x) scales. These models represent state-of-the-art single-stage detectors optimized for real-time performance. Earlier YOLO variants rely on anchor-based detection, while more recent versions (YOLOv8 onwards) adopt anchor-free formulations that directly regress object locations relative to feature map points. Across versions, YOLO models employ multi-scale feature pyramids and decoupled heads for bounding box regression and classification [16].

**Transformer-based Detectors**. To contrast CNN-based architectures, we include RT-DETR, a transformer-based real-time detector that formulates object detection as a set prediction problem. RT-DETR leverages self-attention mechanisms to model global context and removes the need for handcrafted components such as anchor boxes and non-maximal suppression [17]. Both large (l) and extra-large (x) variants are evaluated to assess scalability and performance trade-offs relative to CNN-based detectors.

## 4.5. Ordinal-Aware Classification Loss

Standard object detection frameworks treat class labels as categorical and independent, typically optimizing classification performance using one-hot targets and binary cross-entropy (BCE) loss. However, post-disaster damage assessment inherently involves ordinal labels, where damage states follow a natural progression from undamaged to complete failure. In this setting, not all misclassifications are equally severe. For example, predicting DS2 instead of DS3 represents a minor error compared to predicting DS0 for a DS3 structure.

To better align model supervision with this ordinal structure, we introduce an ordinal-aware classification strategy that modifies the target distribution used in the detection loss while preserving compatibility with existing detection frameworks.

### 4.5.1. Soft Ordinal Target Formulation

We denote the ground-truth damage class for a detected object as $c$, where $c \in \{0, 1, \ldots, C-1\}$ and $C = 5$ corresponds to the five IN-CORE damage states (DS0 through DS4). In standard YOLO training, the detector's task-aligned assigner first matches predicted anchors to ground-truth objects, then assigns each positive match a supervision strength $s \in (0,1]$, based on the alignment quality between predictions and ground truth. This alignment score combines both spatial and semantic matching and is computed as:

$$s = \text{IoU}(b_{\text{pred}}, b_{\text{gt}})^{\beta} \times p_c^{\alpha} \qquad (1)$$

where $\text{IoU}(b_{\text{pred}}, b_{\text{gt}})$ measures the spatial overlap ('Intersection over Union') between the predicted bounding box $b_{\text{pred}}$ and ground-truth box $b_{\text{gt}}$, $p_c$ is the predicted classification confidence for the true class c, and $\alpha$ and β are weighting parameters that control the influence of the classification and localization components respectively (typically $\alpha = 1.0$, β = 6.0) [41]. This metric ensures that anchors with both accurate localization and confident classification receive stronger supervision signals. In the standard formulation, this scalar weight $s$ is applied to a one-hot encoded target vector, concentrating all supervision on class $c$ with zero weight assigned to all other classes i.e. a $R^C$ vector where the $c^{th}$ entry is 1 and all others are 0.

**Gaussian Soft Weighting.** Instead of assigning the full target mass to a single class, we distribute this confidence across neighboring ordinal classes using a Gaussian kernel centered at the true class:

$$w_k = exp\left(-\frac{(k-c)^2}{2\psi^2}\right), \qquad k \in \{0, ..., C-1\} \tag{2}$$

where $k$ indexes the damage classes, $(k - c)$ represents the ordinal distance from the true class, and $\psi$ controls the width of the distribution. Larger values of $\psi$ spread supervision more broadly across distant classes, while smaller values concentrate it near the true class. For example, if the true class is DS2 ($c = 2$) and $\psi = 0.5$, the weights would be approximately [0.0003, 0.135, 1.00, 0.135, 0.0003] for classes DS0 through DS4, assigning highest weight to DS2 with smooth decay toward adjacent classes. However, if $\psi = 1.0$, the weights would be approximately [0.02, 0.36, 1.00, 0.36, 0.02] ensuring weights are spread out more broadly.

**K-Neighbor Truncation.** To prevent excessive smoothing and avoid rewarding large ordinal errors, the distribution is truncated to a bounded neighborhood of radius $K$:

$$w_k = 0 \quad if \ |k-c| > K, \tag{3}$$

For instance, with $K = 1$, only the true class and its immediate neighbors receive non-zero weights. If $c = 2$ (DS2), then only classes $\{1, 2, 3\}$ (DS1, DS2, DS3) would have non-zero weights, while DS0 and DS4 are explicitly zeroed (Figure 3). This bounded truncation preserves ordinal locality and prevents the model from being rewarded for predictions that are far from the ground truth in ordinal space.

**Normalized Soft Targets.** The truncated weights are normalized to sum to one and scaled by the task-aligned confidence $s$, yielding the final soft ordinal target distribution:

$$\tilde{y}_k = s \cdot \frac{w_k}{\sum_{j=0}^{C-1} w_j} \tag{4}$$

where the normalization term $\sum_{j=0}^{C-1} w_j$ ensures that the weights form a valid probability distribution over the non-zero classes, and multiplication by $s$ preserves the detector's task-aligned assignment strength. These soft targets $\tilde{y}_k$ replace the scaled one-hot vectors in the classification loss, providing graded supervision that reflects the ordinal structure of damage severity while maintaining full compatibility with YOLO's task-aligned learning framework.

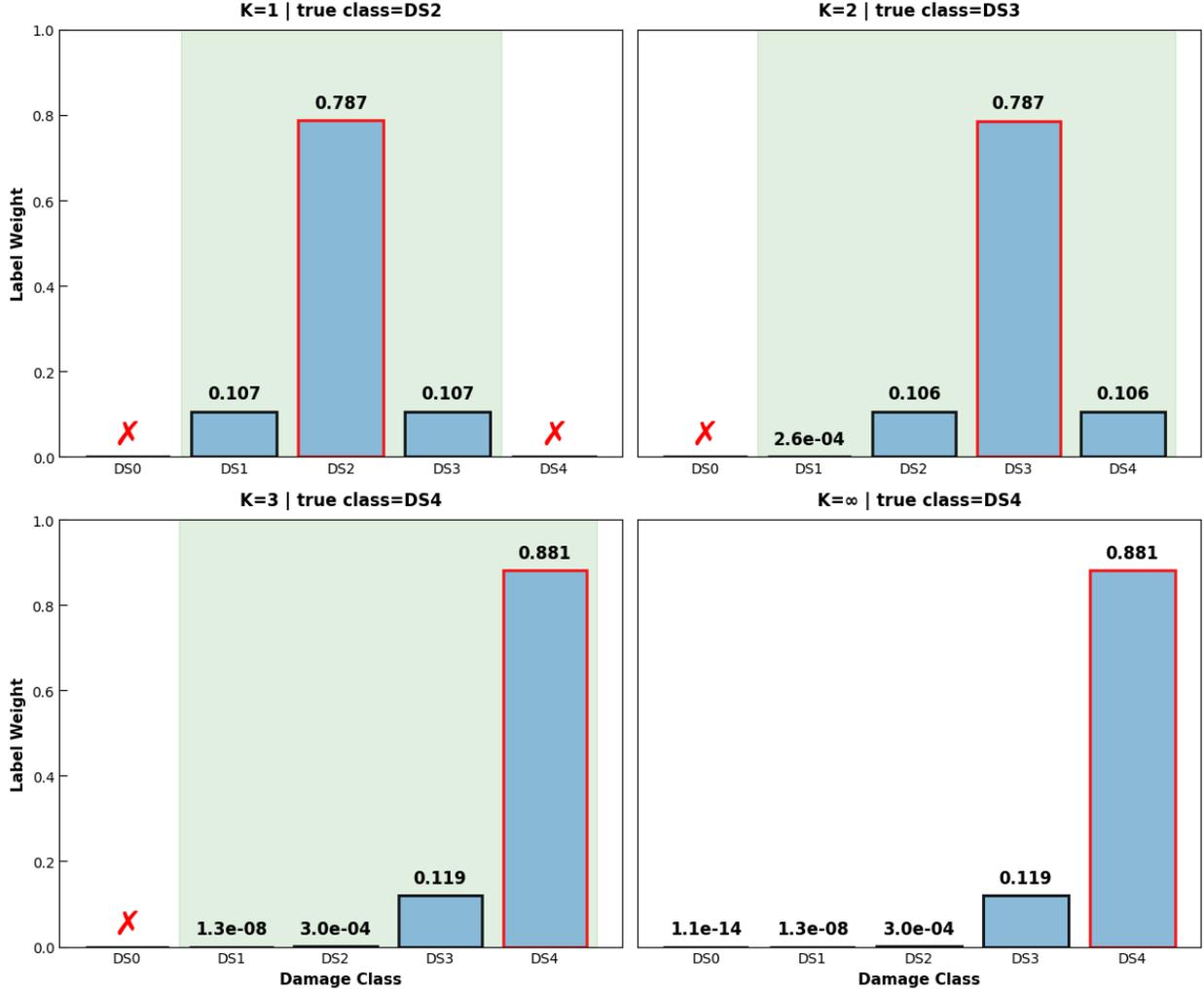

**Figure 3**: Effect of K-neighbor bounding on soft ordinal label distributions. Plots show how increasing $K$ from 1 to ∞ progressively includes more ordinal neighbors in the soft target ($\psi = 0.5$). Red '×' marks indicate zeroed classes while the red outlines show the true class. Green background shaded regions indicate non-zero label support.

### 4.5.2. Ordinal-Aware Classification Loss

**YOLO Classification Loss.** For YOLO-based detectors, given predicted logits $z_k$ for each damage class, classification is optimized using binary cross-entropy (BCE) with logits:

$$\mathcal{L}_{\text{cls}}^{YOLO} = \sum_{k=0}^{C-1} \left[ -\tilde{y}_k \log \sigma(z_k) - (1 - \tilde{y}_k) \log(1 - \sigma(z_k)) \right] \tag{5}$$

where $\sigma(\cdot)$ denotes the sigmoid function and $\tilde{y}_k$ are the soft ordinal targets from Equation 4. This formulation remains fully compatible with multi-label classification heads used in YOLO-style detectors and does not require architectural modification.

**RT-DETR Classification Loss**. RT-DETR employs focal loss, which applies a modulating factor $(1 - p_k)^\gamma$ to down-weight well-classified examples and uses softmax normalization to produce mutually exclusive class probabilities [42]. The focal loss with soft ordinal targets is formulated as:

$$\mathcal{L}_{cls}^{RT-DETR} = -\sum_{k=0}^{C-1} \tilde{y}_k (1 - p_k)^\gamma \log(p_k) \tag{6}$$

where $p_k = \frac{\exp(z_k)}{\sum_{j=0}^{C-1} \exp(z_j)}$ are the softmax-normalized class probabilities, and $\gamma$ is the focusing parameter (typically $\gamma = 1.5$) [41]. For RT-DETR, the predicted class is determined by selecting the class with highest softmax probability, whereas YOLO uses independent sigmoid activations for each class. This preserves the ordinal supervision principle while respecting each detector's native classification formulation.

### 4.5.3. Ordinal Distance Penalty

In addition to soft ordinal targets, we investigate an explicit ordinal-distance penalty designed to further discourage large deviations between predicted and true damage class. For each positive detection, we compute the predicted class index as the weighted average (i.e. expected value) of class probabilities:

$$\hat{c} = \sum_{k=0}^{C-1} k \cdot p_k \tag{7}$$

where $p_k$ is obtained via softmax normalization over class logits, providing a continuous measure of the predicted damage severity. The classification loss for each detection is then scaled by a distance-dependent factor:

For YOLO:

$$\mathcal{L}_{ord}^{YOLO} = (1 + \lambda \cdot |\hat{c} - c|)\mathcal{L}_{cls}^{YOLO} \tag{8}$$

For RT-DETR:

$$\mathcal{L}_{ord}^{RT-DETR} = (1 + \lambda \cdot |\hat{c} - c|)\mathcal{L}_{cls}^{YOLO} \tag{9}$$

where $\lambda \geq 0$ controls the strength of the penalty, and $c$ is the true damage class. This multiplicative factor increases the loss for predictions that are farther from the true damage state while leaving near-miss predictions (e.g., DS2 predicted as DS1) relatively less penalized compared to large errors (e.g., DS2 predicted as DS0).

As shown in Section 5, while soft ordinal targets improve ordinal prediction quality for RT-DETR, the explicit distance penalty can severely degrade detection performance under certain conditions, highlighting a trade-off between ordinal sensitivity and detection stability.

The proposed ordinal-aware loss framework introduces continuous, proximity-based supervision that reflects the ordered nature of damage severity while preserving the strengths of modern object detection architectures. By decoupling ordinal supervision from architectural design, this approach enables a controlled investigation of how loss formulation and model choice jointly influence damage detection performance.

## 5. Experimental Setup

All models were trained and evaluated using a standardized experimental protocol to ensure rigorous comparison across architectures and supervision strategies.

**Training Protocol**. Models were trained at a fixed input resolution of 896 × 896 pixels with a batch size of 16 for a maximum of 250 epochs. To prevent overfitting, an early stopping mechanism was employed with a patience of 10 epochs. All experiments were executed on NVIDIA A100 GPUs (40 GB and 80 GB variants). To ensure statistical reliability, all experiments were evaluated across three random seeds, and multiscale training was enabled to improve model robustness to structural scale variations.

**Detection and Efficiency Metrics**. Primary detection performance was assessed using standard COCO metrics [43], specifically mean Average Precision(mAP) at an IoU (Intersection over Union) threshold of 0.5 ($mAP_{50}$) and the $F1$ score, which is the harmonic mean of precision and recall. To evaluate the feasibility of real-time deployment, we reported inference latency in frames per second (FPS) and model size (parameters in millions and disk size in MB). FPS measurements reflect end-to-end inference time including preprocessing (image resizing and normalization), forward pass, and post-processing (non-maximum suppression), measured on NVIDIA A100 GPUs with batch size 1.

**Ordinal-Aware Metrics**. To explicitly evaluate ordinal damage recognition, ordinal-aware metrics were computed on matched detections. These include the Mean Absolute Ordinal Error (MAOE), which quantifies the average ordinal distance (in class indices) between class index $\hat{c}_i$ and the ground truth index $c_i$:

$$\text{MAOE} = \frac{1}{N} \sum_{i=1}^{N} |\hat{c}_i - c_i| \tag{10}$$

where $N$ is the total number of matched detections, $\hat{c}_i$ is the predicted class index (obtained via $\arg\max$ over predicted class probabilities), and $c_i$ is the true class index. Lower MAOE values indicate better ordinal consistency.

We also computed Ordinal Top-k Accuracy, which measures the percentage of predictions where the predicted damage class is within $k$ ordinal level of the ground truth:

$$\text{Acc}_{\text{ord}}^{(k)} = \frac{1}{N} \sum_{i=1}^{N} \{\mathbb{1}\,(\,|\hat{c}_i - c_i| \leq k\,) \tag{11}$$

where $\mathbb{1}(\cdot)$ is the indicator function that returns 1 if the condition is met or 0 otherwise. For this study, we primarily focus on Ordinal Top-1 Accuracy ($k = 1$), which captures the fraction of predictions that are either exactly correct or are within one damage class of the ground truth.

These metrics are critical for disaster management, as they distinguish minor "adjacent-class" confusion from catastrophic misclassifications (e.g., misidentifying $DS4$ as $DS0$).

## 6. Results and Analysis

### 6.1. Baseline Detection Performance Across Architectures

We first evaluate the baseline performance of all models trained using standard binary cross-entropy classification without ordinal-aware supervision, i.e., $\lambda = 0$ (Equations 8-9) and $K = 0$ (Equation 3), see Table 3. This establishes a common reference point for comparing architectural behavior and for assessing the impact of ordinal loss formulations in subsequent experiments.

Table 3. Baseline Performance of CNN- and Transformer-Based Detectors on TornadoNet Dataset at $K = 0$ and $\lambda = 0$

| Model | mAP@0.5 (%) [↑] | F1 Score (%) [↑] | Ordinal Top-1 (%) [↑] | MAOE [↓] | FPS | Params (M) |
|---|---|---|---|---|---|---|
| YOLOv8n | 40.98 ± 1.38 | 45.11 ± 1.19 | 84.01 ± 1.35 | 0.78 ± 0.02 | 276 | 3.0 |
| YOLOv8l | 42.09 ± 1.16 | 46.41 ± 0.63 | 84.19 ± 0.69 | 0.78 ± 0.02 | 91 | 43.6 |
| YOLOv8x | 41.84 ± 0.72 | 46.24 ± 0.74 | 83.04 ± 0.34 | 0.81 ± 0.00 | 68 | 68.1 |
| YOLO11n | 41.14 ± 0.45 | 45.73 ± 1.17 | 84.79 ± 0.34 | 0.77 ± 0.01 | 239 | 2.6 |
| YOLO11l | 40.44 ± 1.34 | 44.41 ± 0.55 | 83.75 ± 0.22 | 0.79 ± 0.02 | 96 | 25.3 |
| YOLO11x | **46.05 ± 4.35** | **49.40 ± 3.16** | 85.20 ± 1.99 | 0.76 ± 0.06 | 66 | 56.8 |
| RT-DETR-L | 39.87 ± 1.08 | 44.77 ± 1.20 | **88.13 ± 1.87** | **0.65 ± 0.04** | 78 | 32.0 |
| RT-DETR-X | 35.75 ± 3.91 | 41.54 ± 3.27 | 87.74 ± 1.40 | 0.67 ± 0.04 | 79 | 65.5 |

Across CNN-based detectors, larger-capacity YOLO variants generally achieve higher detection accuracy, with YOLO11x obtaining the highest baseline mAP@0.5 (46.05 ± 4.35%) and F1 score (49.40 ± 3.16%). Smaller models such as YOLO11n (41.14% mAP, 239 FPS, 2.6M parameters) and YOLOv8n (40.98% mAP, 276 FPS, 3.0M parameters) demonstrate competitive accuracy while maintaining substantially higher inference throughput, highlighting a clear accuracy–efficiency trade-off within the YOLO family. Model size scaling from 2.6M (YOLO11n) to 56.8M parameters (YOLO11x) yields a 4.9% mAP improvement at the cost of 3.6× slower inference (239 FPS vs 66 FPS), confirming the expected scaling behavior of modern one-stage detectors under fixed-resolution training.

Transformer-based models exhibit a distinct performance profile. While RT-DETR-l underperforms the strongest YOLO variants in mAP (39.87% vs 46.05%), it achieves the lowest MAOE (0.65 vs 0.76 for YOLO11x) and highest Ordinal Top-1 Accuracy (88.13% vs 85.20%), representing a 14% reduction in ordinal error. This indicates that, even without ordinal-aware supervision, RT-DETR produces damage predictions that are closer in ordinal distance to the ground truth, suggesting improved robustness to near-boundary damage states. The larger RT-DETR-x model shows similar ordinal behavior (MAOE 0.67, Top-

1 87.74%) but with reduced detection accuracy (35.75% mAP), reflecting sensitivity to dataset size and class imbalance.

Overall, these baseline results reveal complementary strengths across architectures. CNN-based YOLO models excel in absolute detection accuracy and efficiency, while transformer-based detectors demonstrate improved ordinal consistency in damage severity prediction. These findings motivate our investigation of explicit ordinal supervision in Section 5.2, where we explore whether ordinal-aware losses can improve CNN ordinal performance while preserving detection accuracy, and whether they provide additional benefits to transformer models that already exhibit strong ordinal behavior.

## 6.2. Ablation Studies
### 6.2.1. Effect of Soft Ordinal Supervision on Detection Performance

This ablation investigates the impact of soft ordinal classification targets on detection performance using YOLOv8l as a representative anchor-free CNN detector and RT-DETR-L as a transformer-based, anchor-free alternative due to their comparative sizes (43.6M vs 32.0M respectively). The Gaussian smoothing parameter $\psi$ controls the width of the soft label distribution, thereby determining how much neighboring damage classes contribute to the supervision signal (Table 4 and 5). We varied $\psi \in \{0.1, 0.3, 0.5, 1.0, 2.0\}$ and explored the effect of K-neighbor bounding ($K \in \{1, 2\}$), which restricts soft labels to classes within K ordinal positions of the true class while keeping $\lambda = 0$. All other training and evaluation settings were held constant across three random seeds to assess architectural sensitivity to ordinal supervision strength.

Table 4: Effect of Gaussian Smoothing with $K = 1$ Neighbor Bounding

| Model | $\psi$ | mAP@0.5 (%) [↑] | ΔmAP@0.5 vs Baseline [↑] | F1 Score (%) [↑] | Ordinal Top-1 (%) [↑] | MAOE [↓] |
|---|---|---|---|---|---|---|
| YOLOv8l | baseline | **42.09 ± 1.16** | — | **46.41 ± 0.63** | 84.19 ± 0.69 | 0.78 ± 0.02 |
| | 0.1 | 41.46 ± 1.20 | −0.6 | 45.77 ± 1.09 | 84.02 ± 0.20 | 0.79 ± 0.00 |
| | 0.3 | 41.42 ± 1.73 | −0.7 | 45.33 ± 1.92 | 83.85 ± 1.35 | 0.79 ± 0.03 |
| | 0.5 | 40.75 ± 1.38 | −1.3 | 45.16 ± 0.94 | 83.35 ± 0.40 | 0.82 ± 0.01 |
| | 1.0 | 29.87 ± 5.86 | −12.2 | 35.11 ± 6.02 | 75.11 ± 0.79 | 1.06 ± 0.02 |
| | 2.0 | 24.91 ± 5.11 | −17.2 | 30.81 ± 8.05 | 74.88 ± 3.67 | 1.07 ± 0.13 |
| RT-DETR-L | baseline | 39.87 ± 1.08 | — | 44.77 ± 1.20 | 88.13 ± 1.87 | 0.65 ± 0.04 |
| | 0.1 | 38.50 ± 1.48 | −1.4 | 39.90 ± 2.93 | 88.67 ± 0.76 | 0.66 ± 0.03 |
| | 0.3 | 43.13 ± 0.11 | +3.3 | 45.31 ± 1.66 | 88.91 ± 1.95 | 0.62 ± 0.04 |
| | 0.5 | **44.70 ± 0.94** | +4.8 | **46.13 ± 1.99** | 91.15 ± 0.93 | 0.56 ± 0.03 |
| | 1.0 | 27.92 ± 14.64 | −11.9 | 22.55 ± 12.13 | 89.98 ± 1.98 | 0.63 ± 0.08 |
| | 2.0 | 18.53 ± 9.88 | −21.3 | 7.17 ± 4.98 | 96.89 ± 3.21 | 0.35 ± 0.25 |

Note: For ΔmAP@0.5, values are reported as percentage points (pp). For MAOE, values closer to zero indicate improvement (↓ lower is better).

Table 5: Effect of K-Neighbor Bounding at $\psi = 0.5$

| Model | K | mAP@0.5 (%) [↑] | ΔmAP@0.5 vs Baseline [↑] | F1 Score (%) [↑] | Ordinal Top-1 (%) [↑] | MAOE [↓] |
|---|---|---|---|---|---|---|
| YOLOv8l | baseline | **42.09 ± 1.16** | — | **46.41 ± 0.63** | **84.19 ± 0.69** | **0.78 ± 0.02** |
|  | 1 | 40.75 ± 1.38 | −1.3 | 45.16 ± 0.94 | 83.35 ± 0.40 | 0.82 ± 0.01 |
|  | 2 | 41.38 ± 1.56 | −0.7 | 45.84 ± 1.35 | 83.89 ± 0.58 | 0.80 ± 0.01 |
| RT-DETR-L | baseline | 39.87 ± 1.08 | — | 44.77 ± 1.20 | 88.13 ± 1.87 | 0.65 ± 0.04 |
|  | 1 | **44.70 ± 0.94** | +4.8 | 46.13 ± 1.99 | **91.15 ± 0.93** | **0.56 ± 0.03** |
|  | 2 | 44.39 ± 0.46 | +4.5 | **46.40 ± 0.21** | 90.20 ± 1.62 | 0.59 ± 0.03 |

Note: $k$ denotes maximum ordinal distance for non-zero soft label weights. Results show RT-DETR-L consistently benefits from ordinal supervision ($K = \{1, 2\}$), while YOLOv8l remains near baseline. (↓ lower is better.)

**YOLOv8l Response to Ordinal Supervision.** For the anchor-free YOLOv8l detector, incorporating soft ordinal targets yielded minimal performance changes when $\psi \leq 0.5$. As shown in Table 4, mAP@0.5 remained stable at approximately 41%, varying by less than 1 percentage point (pp) across $\psi \in \{0.1, 0.3, 0.5\}$, effectively matching baseline performance. Similarly, F1 score and ordinal metrics (Ordinal Top-1 Accuracy ~84%, MAOE ~0.79) showed negligible variation within this range. With $\psi = 0.5$ and varying K-neighbor bounds ($K \in \{1, 2\}$), mAP@0.5 remained within 1.5 percentage points of baseline (40.75–41.38% vs. 42.09% baseline), F1 scores stayed near 45%, and ordinal metrics showed negligible variation (Table 5).

However, increasing $\psi$ beyond 0.5 triggered severe performance degradation: at $\psi = 1.0$, mAP@0.5 dropped to 29.87% (−12.2 pp), F1 score fell to 35.11% (−11.3 pp), and MAOE increased to 1.06. This sharp decline indicates that excessive label smoothing undermines the discriminative capacity in anchor-free CNN detectors, likely because the dense prediction heads at multiple scales receive conflicting gradients when supervision is overly diffused across ordinal neighbors.

**RT-DETR-L Response to Ordinal Supervision**. In contrast, the transformer-based RT-DETR-L architecture exhibited clear benefits from ordinal-aware supervision at moderate smoothing levels. As shown in Figure 4 performance peaked at $\psi = 0.5$, achieving mAP@0.5 = 44.70% (+6.2 pp vs. $\psi = 0.1$), F1 score = 46.13% (+6.2 pp), Ordinal Top-1 Accuracy = 91.15%, and MAOE = 0.56 (−0.10 absolute improvement), representing a 4.8% mAP improvement and 14% MAOE reduction over baseline. Compared to YOLOv8l, RT-DETR-L demonstrated superior ordinal awareness even at conservative smoothing levels ($\psi = 0.1$: 88.67% Ordinal Top-1 vs. 84.02% for YOLOv8l). Interestingly, increasing neighborhood size to $K = 2$ while maintaining $\psi = 0.5$ yielded slightly reduced mAP as compared to $K = 1$ but still improved performance (42.49% mAP@0.5, +2.6 pp vs. baseline), suggesting that tighter bounding ($K = 1$) better preserves class discrimination while allowing sufficient ordinal flexibility (Figure 4). These gains suggest that the RT-DETR-L's global attention mechanism and end-to-end bipartite matching process are better suited to exploit ordered class relationships, enabling smoother decision boundaries between adjacent damage severities.

Both architectures exhibited failure modes at $\psi = 2.0$, though with different characteristics. YOLOv8l experienced gradual degradation (24.91% mAP@0.5, 30.81% F1), while RT-DETR-L collapsed catastrophically (18.53% mAP@0.5, 7.17% F1) despite achieving paradoxically high ordinal accuracy (96.89%). Analysis of ordinal confusion matrices reveals the mechanism: the model learned to predict exclusively DS0 (Undamaged), DS1 (Slight) and DS2 (Moderate) classes, the most frequent categories in the dataset, thereby achieving high ordinal accuracy (predictions are rarely >1 class away from ground truth) while completely failing to detect moderate-to-complete damage (see ordinal confusion matrices in Figure 5). This degenerate solution highlights that excessive smoothing can lead models to exploit dataset class imbalance rather than learn meaningful ordinal relationships.

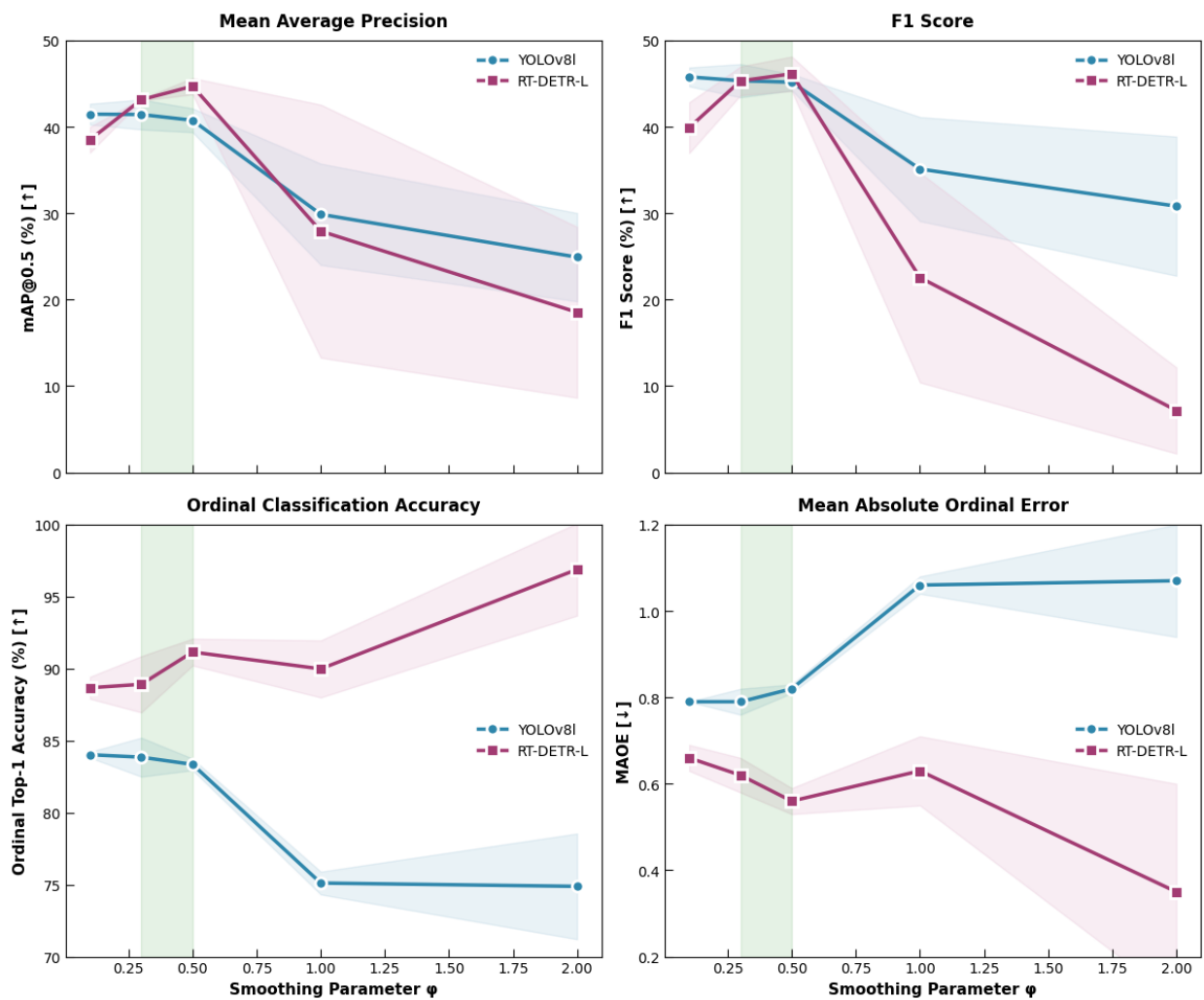

**Figure 4**: Impact of Gaussian Smoothing Parameter $\psi$ on Detection and Ordinal Performance. Shaded regions along line plots represent ± 1 standard deviation across 3 random seeds. Green shaded area ($\psi = 0.3 - 0.5$) indicates optimal smoothing range for RT-DETR-L, yielding +6.2 pp mAP@0.5 improvement over conservative smoothing ($\psi = 0.1$) and 14% MAOE reduction over baseline. YOLOv8l shows negligible benefit from ordinal supervision across all $\psi \leq 0.5$. Both architectures experience severe degradation at $\psi \geq 1.0$ and (↓) means lower is better.

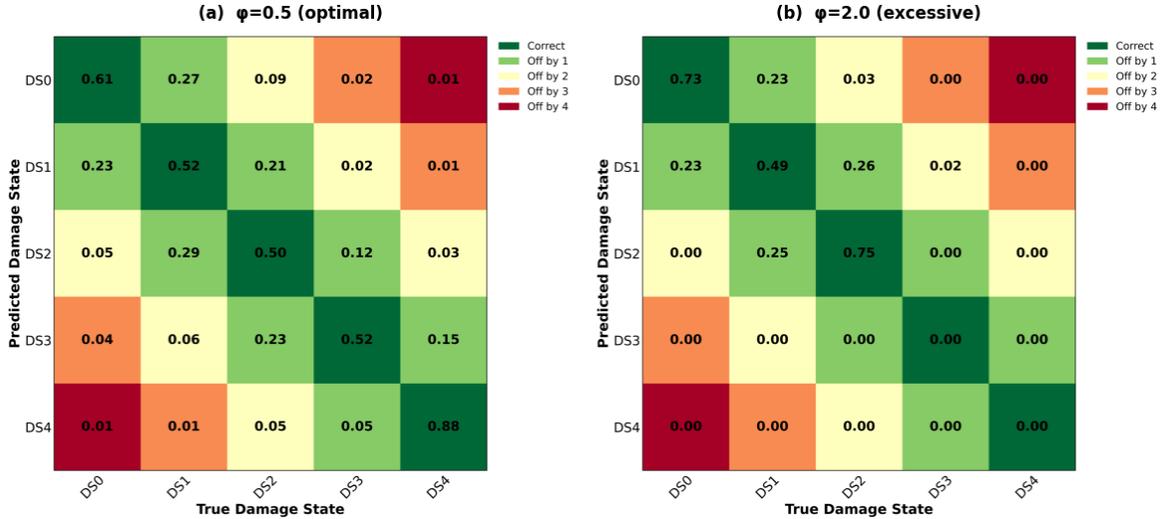

**Figure 5**: Effect of label smoothing strength on prediction behavior. Normalized ordinal confusion matrices (a) at optimal smoothing ($\psi = 0.5$), RT-DETR-L produces well-calibrated predictions across all damage severities. (b) Excessive smoothing ($\psi = 2.0$) causes the model to collapse toward frequent classes (DS0-DS2).

Figure 6 qualitatively compares baseline (a) and ordinal-supervised RT-DETR-L (b) predictions on identical post-tornado scenes. The 2×2 grid format shows ground truth (top-left), true positives (top-right), false positives (bottom-left), and false negatives (bottom-right) for two damage scenarios. In the first scene (left column), the baseline model systematically underestimates damage severity, misclassifying complete damage (DS4) as extensive (DS3) and extensive damage (DS3) as moderate (DS2) across all buildings. In contrast, the ordinal-supervised model correctly identifies the majority of structures, with only a single off-by-one error (DS3 instead of DS4), demonstrating substantially improved severity alignment.

In the second scene (right column), the baseline model confuses moderate damage (DS2) with slight damage (DS1), while the ordinal-supervised model correctly predicts the ground-truth damage state for all four buildings in this scene. Across both examples, ordinal supervision consistently maintains minimal severity jumps and constrains errors to adjacent damage levels when misclassifications occur. These visual results directly corroborate the quantitative findings reported in Table 4, where ordinal supervision improves Ordinal Top-1 Accuracy from 84.19% to 91.15% and reduces MAOE from 0.65 to 0.56 for RT-DETR-L.

**Architectural Implications.** The divergent responses to ordinal supervision highlight fundamental architectural differences. YOLOv8l's anchor-free design with coupled detection and classification heads across multiple feature pyramid levels appears incompatible with soft ordinal supervision. The spatial redundancy in predictions may amplify the confusion introduced by label smoothing. Conversely, RT-DETR-L's sparse set of learned object queries and global class assignment via Hungarian matching naturally accommodates soft, order-aware supervision, as each query receives a single, unified target distribution.

Based on these results, we recommend $\psi = 0.3 - 0.5$ for RT-DETR-L when ordinal supervision is desired, yielding meaningful improvements in both detection and ordinal metrics. For YOLOv8l and similar anchor-free detectors, standard one-hot supervision ($\psi \to 0$ or baseline) remains preferable, as ordinal smoothing provides negligible benefit while risking performance degradation. Across both architectures, $\psi \geq 1.0$ should be avoided, as excessive smoothing consistently degrades performance regardless of detector design.

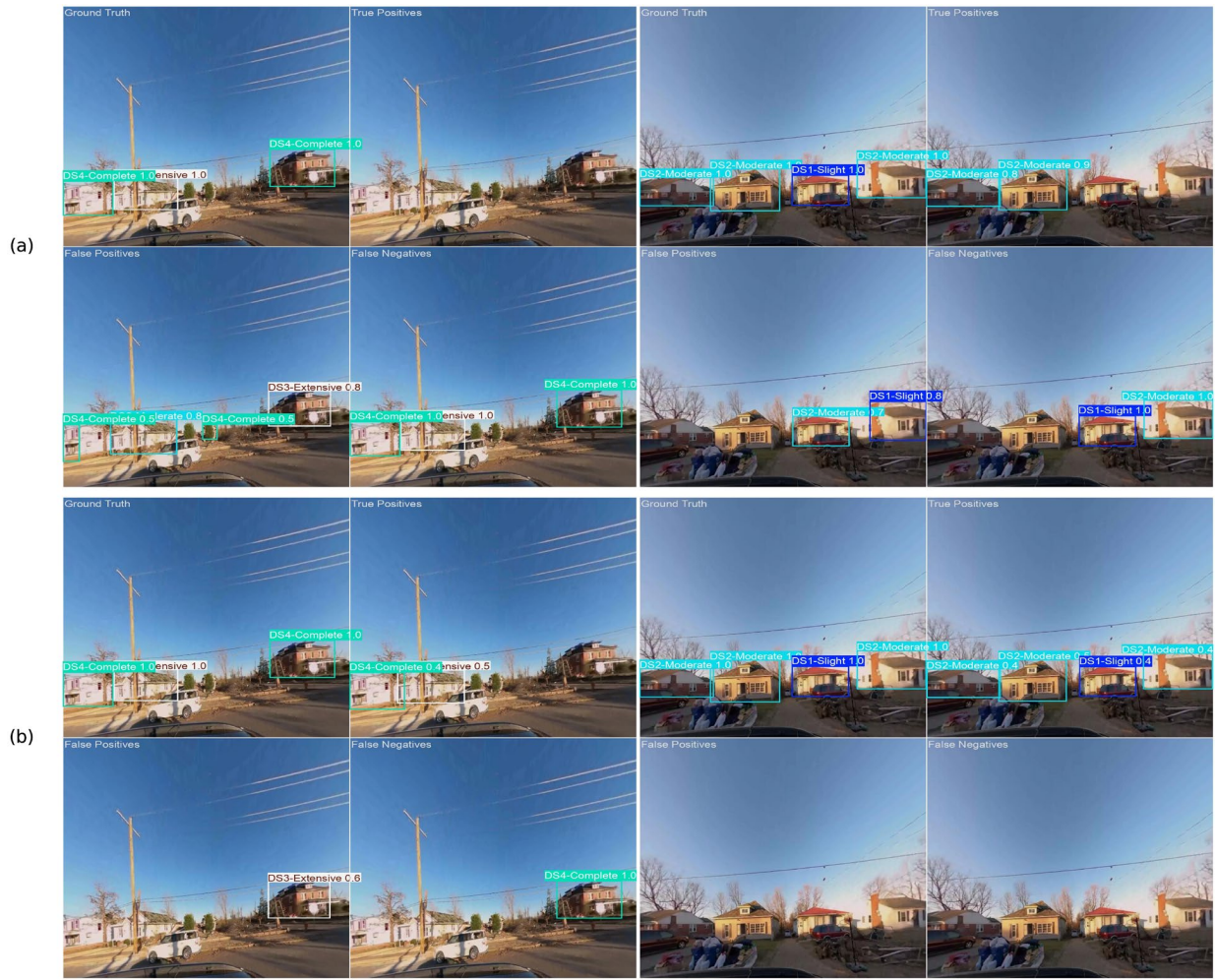

**Figure 6:** Qualitative comparison of baseline and ordinal-supervised RT-DETR-L predictions on identical street-view tornado damage scenes. Two representative scenes are shown across columns. For each scene, the top row (a) shows baseline predictions, and the bottom row (b) shows ordinal-supervised predictions ($\psi = 0.5, K = 1$). Each scene is organized as a 2×2 grid: ground-truth annotations (top-left), true positive detections (top-right), false positives (bottom-left), and false negatives (bottom-right). Bounding boxes are color-coded by predicted damage class.

### 6.2.2. Effect of Explicit Ordinal-Distance Penalty on Detection Performance

We further evaluate whether explicitly penalizing ordinal distance between predicted and ground-truth damage states provides additional benefit beyond standard classification supervision. We consider a distance-based penalty applied to the classification loss, parameterized by a scalar weight $\lambda$, with no ordinal smoothing applied to the targets ($K = 0$). Experiments are conducted on the same representative architectures from the previous ablation study: YOLOv8-l and RT-DETR-L.

Results in Table 6 show that ordinal penalty ($\lambda \in \{0.05, 0.1, 0.2\}$) consistently improves RT-DETR-L performance, with $\lambda = 0.05$ achieving 43.36% mAP@0.5 (+3.5 pp vs. baseline) and $\lambda = 0.20$ reaching 43.24% mAP@0.5 (+3.4 pp) with slightly better ordinal metrics (MAOE=0.59). In contrast, YOLOv8l remains at baseline performance (~42% mAP@0.5) across all $\lambda$ values, further confirming that anchor-free CNNs do not benefit from ordinal-aware supervision regardless of implementation strategy.

Comparing the two ordinal approaches: Gaussian label smoothing ($\psi = 0.5$, $K = 1$: 44.70% mAP) outperforms ordinal penalty (best: 43.36% mAP) for RT-DETR-L, suggesting that soft target distributions are more effective than loss-based penalties for transformer architectures with focal loss. However, ordinal penalty offers a simpler implementation requiring no target modification, making it a practical alternative when label smoothing is undesirable.

Table 6: Effect of Ordinal Penalty Parameter $\lambda$ ($K = 0$)

| Model | $\lambda$ | mAP@0.5 (%) [↑] | ΔmAP@0.5 vs Baseline [↑] | F1 Score (%) [↑] | Ordinal Top-1 (%) [↑] | MAOE [↓] |
|---|---|---|---|---|---|---|
| YOLOv8l | baseline | 42.09 ± 1.16 | — | **46.41 ± 0.63** | **84.19 ± 0.69** | 0.78 ± 0.02 |
| | 0.05 | 42.08 ± 1.99 | 0.0 | 45.99 ± 1.65 | 84.02 ± 0.20 | **0.77 ± 0.02** |
| | 0.1 | **42.66 ± 1.40** | **+0.6** | 46.14 ± 1.42 | 83.85 ± 1.35 | 0.81 ± 0.01 |
| | 0.2 | 42.13 ± 0.76 | 0.0 | 45.94 ± 0.89 | 83.35 ± 0.40 | 0.78 ± 0.03 |
| RT-DETR-L | baseline | 39.87 ± 1.08 | — | 44.77 ± 1.20 | 88.13 ± 1.87 | 0.65 ± 0.04 |
| | 0.05 | **43.36 ± 0.67** | **+3.5** | **45.73 ± 0.58** | 89.54 ± 0.29 | 0.61 ± 0.01 |
| | 0.1 | 41.85 ± 1.15 | +2.0 | 42.36 ± 1.08 | 89.93 ± 1.61 | 0.61 ± 0.04 |
| | 0.2 | 43.24 ± 0.87 | +3.4 | 45.16 ± 1.54 | **90.73 ± 0.30** | **0.59 ± 0.02** |

## 7. Summary and Conclusions

This study presents TornadoNet, a comprehensive benchmark for automated street-view building damage assessment that advances the methodological understanding of ordinal supervision in object detection and provides deployment-relevant evidence for AI-assisted post-disaster assessment. Using 3,333 high-resolution geotagged images and 8,890 annotated building instances collected following the 2021 Midwest tornado outbreak, we conducted a systematic comparison of modern CNN-based and transformer-based real-time detectors for multi-level damage classification under standardized training and evaluation protocols.

Baseline experiments revealed complementary strengths across architectures. CNN-based YOLO models achieved the highest detection accuracy, with YOLO11x obtaining 46.05% mAP@0.5 and 49.40% F1 score while maintaining real-time inference speeds. Smaller variants such as YOLO11n (41.14% mAP@0.5, 239 FPS) delivered competitive accuracy with substantially higher throughput, illustrating clear efficiency–accuracy trade-offs. Transformer-based RT-DETR-L, while achieving lower baseline mAP@0.5 (39.87%), exhibited superior ordinal consistency with 88.13% Ordinal Top-1 Accuracy and MAOE of 0.65 compared to 85.20% and 0.76 for the best YOLO variant, corresponding to a meaningful reduction in average ordinal error even without ordinal-aware supervision.

To better align predictions with the ordered nature of structural damage, we introduced soft ordinal classification targets and explicit ordinal distance penalties as alternatives to standard one-hot supervision. Results show a clear architecture-dependent response to ordinal-aware loss formulations. RT-DETR-L benefited from soft ordinal supervision when the smoothing strength was carefully calibrated. With Gaussian smoothing ($\psi = 0.5$, $K = 1$), RT-DETR-L reached 44.70% mAP@0.5, improving by 4.8 percentage points over baseline, while also improving ordinal metrics (91.15% Ordinal Top-1 Accuracy, MAOE = 0.56). Error patterns indicate that ordinal supervision tends to concentrate mistakes among adjacent damage states, reducing far-off confusion that is most problematic for severity grading. In contrast, the anchor-free CNN detector YOLOv8l showed limited sensitivity to ordinal supervision, with performance remaining near baseline across configurations tested. This divergence is consistent with differences in training dynamics and assignment, where RT-DETR's sparse object queries and one-to-one matching can exploit structured targets more directly than spatially redundant dense predictions. The explicit ordinal distance penalty produced similar trends, improving RT-DETR-L while providing no consistent benefit to YOLOv8l, reinforcing that ordinal-aware supervision interacts strongly with detector design.

From an efficiency standpoint, all evaluated models achieved real-time throughput on A100 GPUs, supporting their use in time-constrained analysis pipelines. For scenarios that prioritize computational efficiency, smaller YOLO variants provide strong speed with competitive detection accuracy. For scenarios where ordinal consistency is a primary requirement, RT-DETR-L with calibrated ordinal supervision provides stronger severity coherence while maintaining acceptable throughput.

Beyond methodological advances in object detection, the implications of this work extend directly to the practice of post-disaster damage evaluation and resilience planning. Rapid, standardized, and building-level damage assessment forms the foundation for emergency response prioritization, resource allocation, insurance estimation, and long-term recovery planning. By demonstrating that real-time AI systems can achieve competitive detection accuracy while preserving ordinal severity coherence, TornadoNet represents a step toward accurate automation of damage surveys that traditionally require extensive field personnel and time-intensive manual inspection. Automated street-level assessment frameworks have the potential to supplement reconnaissance teams, reduce exposure to hazardous environments, and provide consistent, reproducible severity grading across large geographic areas. Over time, the availability of systematically collected, fine-grained damage data can also inform vulnerability modeling, calibration of fragility curves, and refinement of building codes and performance-based design standards. In this way, advances in ordinal-aware detection do not merely improve algorithmic performance but contribute to a broader infrastructure resilience ecosystem by enabling faster feedback loops between hazard events, observed structural performance, and engineering design improvements.

This work contributes (1) a controlled benchmark of CNN- and transformer-based detectors for street-view tornado damage assessment, (2) ordinal-aware loss formulations and evaluation metrics for multi-level damage detection, (3) empirical evidence that the effectiveness of ordinal supervision is architecture-dependent, and (4) the release of TornadoNet trained models, training code, and datasets to support reproducibility and future research (**Model & Data**: https://github.com/crumeike/TornadoNet). Future work should evaluate generalization across events and regions, investigate stronger imbalance mitigation for rare severe damage classes, and study hybrid approaches that combine CNN efficiency with transformer-based ordinal consistency under larger and more diverse datasets.

## 8. Acknowledgments

The authors acknowledge the Center for Risk-Based Community Resilience Planning, a NIST-funded Center of Excellence (Cooperative Agreement 70NANB15H044), for providing the street-view imagery dataset, and SciServer (sciserver.org), developed at Johns Hopkins University and funded by NSF Award ACI-1261715, for computational resources. These resources were instrumental in the development and validation of the models presented in this work.

## 9. Disclaimer

Certain equipment, instruments, software, or materials are identified in this paper in order to specify the experimental procedure adequately. Such identification is not intended to imply recommendation or endorsement of any product or service by NIST, nor is it intended to imply that the materials or equipment identified are necessarily the best available for the purpose.

## References

[1] Prevatt DO, van de Lindt JW, Back EW, Graettinger AJ, Pei S, Coulbourne W, et al. Making the Case for Improved Structural Design: Tornado Outbreaks of 2011. Leadership and Management in Engineering 2012;12:254–70. https://doi.org/10.1061/(ASCE)LM.1943-5630.0000192.

[2] Meng Z, Yao D. Damage Survey, Radar, and Environment Analyses on the First-Ever Documented Tornado in Beijing during the Heavy Rainfall Event of 21 July 2012. Weather Forecast 2014;29:702–24. https://doi.org/10.1175/WAF-D-13-00052.1.

[3] Johnston B, Wang L, van de Lindt JW, Harati M, Skakel K, Crawford S, et al. Interdisciplinary data collection for empirical community-level recovery modelling, 2024, p. 1260–7. https://doi.org/10.2749/manchester.2024.1260.

[4] Burgess D, Ortega K, Stumpf G, Garfield G, Karstens C, Meyer T, et al. 20 May 2013 Moore, Oklahoma, Tornado: Damage Survey and Analysis. Weather Forecast 2014;29:1229–37. https://doi.org/10.1175/WAF-D-14-00039.1.

[5] Púčik T, Rýva D, Staněk M, Šinger M, Groenemeijer P, Pistotnik G, et al. The Violent Tornado on 24 June 2021 in Czechia: Damage Survey, Societal Impacts, and Lessons Learned. Weather, Climate, and Society 2024;16:411–29. https://doi.org/10.1175/WCAS-D-23-0080.1.

[6] Marshall TP. Tornado Damage Survey at Moore, Oklahoma. Weather Forecast 2002;17:582–98. https://doi.org/10.1175/1520-0434(2002)017<0582:TDSAMO>2.0.CO;2.

[7] Roueche DB, Prevatt DO. Residential Damage Patterns Following the 2011 Tuscaloosa, AL and Joplin, MO Tornadoes. Journal of Disaster Research 2013;8:1061–7. https://doi.org/10.20965/jdr.2013.p1061.

[8] Kuligowski ED, Lombardo FT, Phan LT, Levitan ML, Jorgensen DP. Draft final report, National Institute of Standards and Technology (NIST) Technical investigation of the May 22, 2011 tornado in Joplin, Missouri. Gaithersburg, MD: 2013. https://doi.org/10.6028/NIST.NCSTAR.3.

[9] Kuligowski ED, Phan LT, Levitan ML, Jorgensen DP. Preliminary Reconnaissance of the May 20, 2013, Newcastle-Moore Tornado in Oklahoma. Gaithersburg, MD: 2013. https://doi.org/10.6028/NIST.SP.1164.

[10] Kijewski-Correa T, Roueche DB, Mosalam KM, Prevatt DO, Robertson I. StEER: A Community-Centered Approach to Assessing the Performance of the Built Environment after Natural Hazard Events. Front Built Environ 2021;7. https://doi.org/10.3389/fbuil.2021.636197.

[11] Rathje EM, Dawson C, Padgett JE, Pinelli J-P, Stanzione D, Adair A, et al. DesignSafe: New Cyberinfrastructure for Natural Hazards Engineering. Nat Hazards Rev 2017;18. https://doi.org/10.1061/(ASCE)NH.1527-6996.0000246.

[12] Gupta R, Goodman B, Patel N, Hosfelt R, Sajeev S, Heim E, et al. Creating xBD: A Dataset for Assessing Building Damage from Satellite Imagery. Proceedings of the IEEE/CVF Conference on Computer Vision and Pattern Recognition (CVPR) Workshops, 2019.

[13] Rahnemoonfar M, Chowdhury T, Sarkar A, Varshney D, Yari M, Murphy RR. FloodNet: A High Resolution Aerial Imagery Dataset for Post Flood Scene Understanding. IEEE Access 2021;9:89644–54. https://doi.org/10.1109/ACCESS.2021.3090981.

[14] Wang W "Lisa," van de Lindt JW, Johnston B, Crawford PS, Yan G, Dao T, et al. Application of Multidisciplinary Community Resilience Modeling to Reduce Disaster Risk: Building Back Better. Journal of Performance of Constructed Facilities 2024;38. https://doi.org/10.1061/JPCFEV.CFENG-4650.

[15] Liu W, Anguelov D, Erhan D, Szegedy C, Reed S, Fu C-Y, et al. SSD: Single Shot MultiBox Detector, 2016, p. 21–37. https://doi.org/10.1007/978-3-319-46448-0_2.

[16] Redmon J, Divvala S, Girshick R, Farhadi A. You Only Look Once: Unified, Real-Time Object Detection. 2016 IEEE Conference on Computer Vision and Pattern Recognition (CVPR), IEEE; 2016, p. 779–88. https://doi.org/10.1109/CVPR.2016.91.

[17] Zhao Y, Lv W, Xu S, Wei J, Wang G, Dang Q, et al. DETRs Beat YOLOs on Real-time Object Detection. 2024 IEEE/CVF Conference on Computer Vision and Pattern Recognition (CVPR), IEEE; 2024, p. 16965–74. https://doi.org/10.1109/CVPR52733.2024.01605.

[18] Ghosh Mondal T, Jahanshahi MR, Wu R, Wu ZY. Deep learning-based multi-class damage detection for autonomous post-disaster reconnaissance. Struct Control Health Monit 2020;27. https://doi.org/10.1002/stc.2507.

[19] van de Lindt JW, Kruse J, Cox DT, Gardoni P, Lee JS, Padgett J, et al. The interdependent networked community resilience modeling environment (IN-CORE). Resilient Cities and Structures 2023;2:57–66. https://doi.org/10.1016/J.RCNS.2023.07.004.

[20] Umeike R, Dao T, Crawford S. Accelerating Post-Tornado Disaster Assessment Using Advanced Deep Learning Models. 2024 IEEE MetroCon, IEEE; 2024, p. 1–3. https://doi.org/10.1109/MetroCon62511.2024.10883935.

[21] Mutis I, Joshi VA, Singh A. Object Detectors for Construction Resources Using Unmanned Aerial Vehicles. Practice Periodical on Structural Design and Construction 2021;26. https://doi.org/10.1061/(asce)sc.1943-5576.0000598.

[22] Zhu X, Liang J, Hauptmann A. MSNet: A Multilevel Instance Segmentation Network for Natural Disaster Damage Assessment in Aerial Videos. 2021 IEEE Winter Conference on Applications of Computer Vision (WACV), IEEE; 2021, p. 2022–31. https://doi.org/10.1109/WACV48630.2021.00207.

[23] Singh DK, Hoskere V. Post Disaster Damage Assessment Using Ultra-High-Resolution Aerial Imagery with Semi-Supervised Transformers. Sensors 2023;23:8235. https://doi.org/10.3390/s23198235.

[24] Mehta KC. Development of the EF-Scale for Tornado Intensity. Journal of Disaster Research 2013;8:1034–41. https://doi.org/10.20965/jdr.2013.p1034.

[25] Brown TM. Development of a statistical relationship between ground-based and remotely-sensed damage in windstorms. Texas Tech University, 2010.

[26] Scawthorn C, Flores P, Blais N, Seligson H, Tate E, Chang S, et al. HAZUS-MH Flood Loss Estimation Methodology. II. Damage and Loss Assessment. Nat Hazards Rev 2006;7:72–81. https://doi.org/10.1061/(ASCE)1527-6988(2006)7:2(72).

[27] Masoomi H, Ameri MR, van de Lindt JW. Wind Performance Enhancement Strategies for Residential Wood-Frame Buildings. Journal of Performance of Constructed Facilities 2018;32. https://doi.org/10.1061/(ASCE)CF.1943-5509.0001172.

[28] Masoomi H, van de Lindt JW. Tornado fragility and risk assessment of an archetype masonry school building. Eng Struct 2016;128:26–43. https://doi.org/10.1016/J.ENGSTRUCT.2016.09.030.

[29] Memari M, Attary N, Masoomi H, Mahmoud H, van de Lindt JW, Pilkington SF, et al. Minimal Building Fragility Portfolio for Damage Assessment of Communities Subjected to Tornadoes. Journal of Structural Engineering 2018;144. https://doi.org/10.1061/(ASCE)ST.1943-541X.0002047.

[30] Koliou M, van de Lindt JW, McAllister TP, Ellingwood BR, Dillard M, Cutler H. State of the research in community resilience: progress and challenges. Sustain Resilient Infrastruct 2020;5:131–51. https://doi.org/10.1080/23789689.2017.1418547.

[31] Carani S, Pingel TJ. Detection of Tornado damage in forested regions via convolutional neural networks and uncrewed aerial system photogrammetry. Natural Hazards 2023;119:143–66. https://doi.org/10.1007/s11069-023-06125-4.

[32] Kim D, Won J, Lee E, Park KR, Kim J, Park S, et al. Disaster assessment using computer vision and satellite imagery: Applications in detecting water-related building damages. Front Environ Sci 2022;10. https://doi.org/10.3389/fenvs.2022.969758.

[33] Radhika S, Tamura Y, Matsui M. Determination of Degree of Damage on Building Roofs Due to Wind Disaster from Close Range Remote Sensing Images Using Texture Wavelet Analysis. IGARSS 2018 - 2018 IEEE International Geoscience and Remote Sensing Symposium, IEEE; 2018, p. 3366–9. https://doi.org/10.1109/IGARSS.2018.8519282.

[34] Xiong F, Wen H, Zhang C, Song C, Zhou X. Semantic segmentation recognition model for tornado-induced building damage based on satellite images. Journal of Building Engineering 2022;61:105321. https://doi.org/10.1016/j.jobe.2022.105321.

[35]   Braik AM, Koliou M. Post-tornado automated building damage evaluation and recovery prediction by integrating remote sensing, deep learning, and restoration models. Sustain Cities Soc 2025;123:106286. https://doi.org/10.1016/j.scs.2025.106286.

[36]   Chen Z, Wagner M, Das J, Doe RK, Cerveny RS. Data-Driven Approaches for Tornado Damage Estimation with Unpiloted Aerial Systems. Remote Sens (Basel) 2021;13:1669. https://doi.org/10.3390/rs13091669.

[37]   Rahnemoonfar M, Chowdhury T, Murphy R. RescueNet: A High Resolution UAV Semantic Segmentation Dataset for Natural Disaster Damage Assessment. Sci Data 2023;10:913. https://doi.org/10.1038/s41597-023-02799-4.

[38]   Hong Z, Zhong H, Pan H, Liu J, Zhou R, Zhang Y, et al. Classification of Building Damage Using a Novel Convolutional Neural Network Based on Post-Disaster Aerial Images. Sensors 2022;22:5920. https://doi.org/10.3390/s22155920.

[39]   van de Lindt JW, Wang W "Lisa," Johnston B, Crawford PS, Yan G, Dao T, et al. Social Susceptibility–Driven Longitudinal Tornado Reconnaissance Methodology: 2021 Midwest Quad-State Tornado Outbreak. ASCE OPEN: Multidisciplinary Journal of Civil Engineering 2025;3. https://doi.org/10.1061/AOMJAH.AOENG-0065.

[40]   Crawford PS, Al-Zarrad MA, Graettinger AJ, Hainen AM, Back E, Powell L. Rapid Disaster Data Dissemination and Vulnerability Assessment through Synthesis of a Web-Based Extreme Event Viewer and Deep Learning. Advances in Civil Engineering 2018;2018. https://doi.org/10.1155/2018/7258156.

[41]   Jocher G, Qiu J, Chaurasia A. Ultralytics YOLO 2023.

[42]   Lin T-Y, Goyal P, Girshick R, He K, Dollar P. Focal Loss for Dense Object Detection. IEEE Trans Pattern Anal Mach Intell 2020;42:318–27. https://doi.org/10.1109/TPAMI.2018.2858826.

[43]   Lin T-Y, Maire M, Belongie S, Hays J, Perona P, Ramanan D, et al. Microsoft COCO: Common Objects in Context, 2014, p. 740–55. https://doi.org/10.1007/978-3-319-10602-1_48.